# Cloud Ensemble Learning for Fault Diagnosis of Rolling Bearings with Stochastic Configuration Networks

Wei Dai [a,b], Jiang Liu [b], Lanhao Wang [c,*]

[a] Artificial Intelligence Research Institute, China University of Mining and Technology, Xuzhou 221116, China
[b] School of Information and Control Engineering, China University of Mining and Technology, Xuzhou 221116, China
[c] National Engineering Research Center of Coal Preparation and Purification, China University of Mining and Technology, Xuzhou 221116, China

ABSTRACT

Fault diagnosis of rolling bearings is of great significance for post-maintenance in rotating machinery, but it is a challenging work to diagnose faults efficiently with a few samples. Additionally, faults commonly occur with randomness and fuzziness due to the complexity of the external environment and the structure of rolling bearings, hindering effective mining of fault characteristics and eventually restricting accuracy of fault diagnosis. To overcome these problems, stochastic configuration network (SCN) based cloud ensemble learning, called SCN-CEL, is developed in this work. Concretely, a cloud feature extraction method is first developed by using a backward cloud generator of normal cloud model to mine the uncertainty of fault information. Then, a cloud sampling method, which generates enough cloud droplets using bidirectional cloud generator, is proposed to extend the cloud feature samples. Finally, an ensemble model with SCNs is developed to comprehensively characterize the uncertainty of fault information and advance the generalization performance of fault diagnosis machine. Experimental results demonstrate that the proposed method indeed performs favorably for distinguishing fault categories of rolling bearings in the few shot scenarios.

## 1. Introduction

As a crucial part of rotating machinery in modern industry [1], rolling bearings cloud not only support the stable operation of the parts between the shafts during rotation conditions, but also make the shafts in the normal working position and improve the accuracy of rotation in the process of working. However, as service time grows, rolling bearings are extremely prone to all kinds of failures due to a variety of uncertain factors such as external temperature, dust, humidity, vibration, etc. Obviously, its failures will lead to serious problems in production efficiency and safety. Therefore, accurate fault identification for the rolling bearings is of great significance to formulate effective maintenance strategies, improve utilization and ensure stable operation of equipment.

Vibration signal analysis methods are widely used for fault diagnosis of rolling bearings. The mainstream methods usually contain two components: feature extraction and fault classification [2]. By analyzing the original vibration signals effectively and providing accurate diagnosis results automatically, vibration signal analysis methods achieve gratifying results. Among the vibration signal analysis methods, time-frequency analysis can effectively process the frequency components in the time domain and the time series arrangement of each frequency band. Besides, many advanced signal processing techniques such as wavelet transform [3], empirical mode decomposition (EMD) [4], variational mode decomposition (VMD) [5], etc., have been investigated for fault diagnosis of rolling bearings.

Although these methods have promoted the improvement in fault diagnosis, they are often restricted by the non-stationary and nonlinear characteristics of fault signals. With the development of nonlinear dynamic analysis theory, many nonlinear parameter estimation methods provide good alternatives for extracting the nonlinear and non-stationary fault features hidden in the vibration signals. Distinct derivative entropies have been reported to extract fault feature signals such as appropriate entropy (ApEn) [6], sample entropy (SampEn) [7], permutation entropy [8], fuzzy entropy [9] and multiscale entropy (MSE) [10]. The above methods, however, do not take into account the feature information uncertainties, which are always inevitable in practice. In practical plants, the uncertainties are always caused by the relatively terrible working environment of rotating machinery, the complex mechanical structure of the equipment, as well as the mechanism of faults with randomness and fuzziness [11]. These factors intensify the fault feature uncertainty, thereby making it difficult to directly use the collected signals to represent fault information accurately. Consequently, the high accuracy of fault diagnosis cannot always be guaranteed by using the existing methods.

___________________
* Corresponding author.
 *E-mail address:* wanglanhao@cumt.edu.cn

In terms of the fault classification, an abundant of machine learning models have been successfully applied and have attracted wide attention in the field of rolling bearing fault diagnosis. The common models contain shallow learner models and deep learner models. The former includs but are not limited to support vector machine (SVM) [12], support vector data description (SVDD) [13], relevance vector machine (RVM) [14], random vector functional-link network (RVFLN) [15], back propagation neural network (BPNN) [16]. Among them, SVM, SVDD and RVM are commonly used for binary classification tasks. Although "One-versus-Rest" or "One-versus-One" can be applied to solve multiple classification problems, they will increase the computational complexity of classifier with the number of categories. For BPNN, the gradient descent-based training algorithms are always susceptible to being trapped in local minima and incur slow convergence, and assigning the appropriate number of network layers and nodes is still an open problem in different tasks. As a randomized learner model, RVFLN has also two crucial issues that make it less practical in fault diagnosis, i.e., (i) manually define the number of hidden nodes; (ii) inefficiently search an appropriate interval for assigning hidden-nodes parameters. When applying the above shallow learner models to practical applications, it is necessary to carry out artificial feature extraction and selection owing to their poor ability.

In contrast, the deep learner models, such as convolutional neural network (CNN) [17-18], recurrent neural network (RNN) [19-20], auto encoder (AE) [21], and deep belief network (DBN) [22], employ end-to-end learning technology to automatically extract features and classify faults. However, there are still some problems, such as time-consuming training and expensive resource occupancy. Besides, the quality of the deep learner models relies heavily on the scale and quality of the samples. There is a risk of serious overfitting when the training dataset is insufficient. Unfortunately, rotating machinery equipment runs normally most of the time and rarely breaks down, which makes it difficult to collect enough fault samples. In practice, the bearing fault diagnosis is a typical few-shot classification issue. Consequently, deep learner model-based bearing fault diagnosis often suffers from low accuracy and limited practical applicability due to their failure to capture effective feature representations and train a large number of model parameters in the few-shot scenarios [23].

To solve the problem of rolling bearing fault diagnosis in the few-shot scenarios, this paper proposes a cloud ensemble learning method with stochastic configuration networks (SCN-CEL). In SCN-CEL, cloud spatial features, expectation, entropy and hyper entropy have been extracted to express the intension of fault concept, which mines not only the randomness, but also the fuzziness of rolling bearing faults. A cloud sampling method is developed to expand the samples in cloud feature domain by using a backward cloud generator and a forward cloud generator. Meanwhile, as a random incremental learner model, SCNs [24] are applied for recognizing health states of bearings. By using SCNs, the fault classification model enables rapid training, effectively reducing the need for artificial parameter adjustment and significantly improving both training efficiency and generalization ability. Besides, a cloud ensemble learning method is proposed to immensely improve the generalization performance of SCNs under small samples. Compared with the traditional ensemble learning based on Bagging that handily leads to over-fitting because of randomly sampling from small samples, the cloud ensemble learning can express randomness of faults as much as possible through multiple cloud models [25]. The main works of this paper are summarized as follows:

- To fully represent randomness and fuzziness of bearings fault information, cloud features are extracted from a limited number of vibration signal samples by using a backward cloud generator without hyper parameter settings and then the low-dimensional, fault-sensitive, and nonlinear fault feature vectors can be obtained.
- A novel sampling method named cloud sampling is developed to overcome the few-shot problem in rolling bearing fault diagnosis by integrating a backward cloud generator and a forward cloud generator. The cloud sampling generates multiple cloud droplets to expand the number of cloud characteristic samples for various faults, enabling the bidirectional transformation from connotation to extension of features information.
- To enhance self-learning ability and generalization performance of classifier under fault small samples, cloud ensemble learning with stochastic configuration networks (SCN-CEL) is developed to automatically and quickly identify the healthy state of rolling bearing. SCN-CEL performs cloud sampling multiple times to obtain cloud samples sets as the input for stochastic configuration networks, which comprehensively reflect the representation of uncertainty information in the cloud model and ensure the diversity of base models.

The rest of this paper is organized as follows. In section 2, the proposed method for bearing fault diagnosis is illustrated in detail. Experiments are carried out to evaluate the proposed method in section 3. Finally, section 4 draws our concluding remarks.

## 2. SCN-CEL for fault diagnosis of rolling bearings

*2.1 Overall framework*

Probability theory [26], fuzzy set [27] and cloud model [28] have been proposed to deal with uncertain information in the rolling bearing system. The selection of membership function in fuzzy theory is subjective and ignores the uncertainty of the membership function itself. The event uncertainty based on probability representation only focuses on the uncertainty of event occurrence and can not describe the semantic uncertainty of events. The cloud model, consisting of a forward cloud generator (FCG) and a backward cloud generator (BCG), is a cognitive model that effectively addresses the problem of uncertainty in bidirectional transformation between the connotation and extension of qualitative concepts, as illustrated in Figure 1. For example, let cloud numerical characteristics (*Ex*, *En*, *He*) express a healthy state of rolling bearing over a period, where *Ex* represents an expected value of vibration signal under certain health state. As different individuals have varying interpretations about that health state with its *Ex*, especially in complex industrial environments, it is challenging to assign a crisp membership degree, while the cloud model can depict this scenario. The normal cloud map of "a healthy state with an *Ex* of vibration signal" is shown in Fig. 1 by FCG. From Fig. 1, we can see that it will get different membership degrees for the same value of vibration signal, and the membership degree also has randomness for the same value. The cloud model allows a stochastic disturbance of the membership degree encircling a determined central value. Corresponding to the FCG, the intension of a healthy state is derived from amount of data, which is a process of acquiring knowledge from quantitative data.

Cloud model, which can explain many common random and fuzzy problems, as well as the relationship between them at the same time, has been successfully applied to many theoretical and practical fields, such as condition assessment [29], risk assessment [30], and fault diagnosis [31]. If the membership function of the cloud model is different, the cloud model will have different distribution shapes, of which the normal cloud model is universal [32]. Accordingly, this research focuses on the normal cloud model. Under the basic assumption that random variables of the cloud model follow the Gaussian distribution, the corresponding relationship between the cloud model and the statistical theory can be established in Table 1 [33].

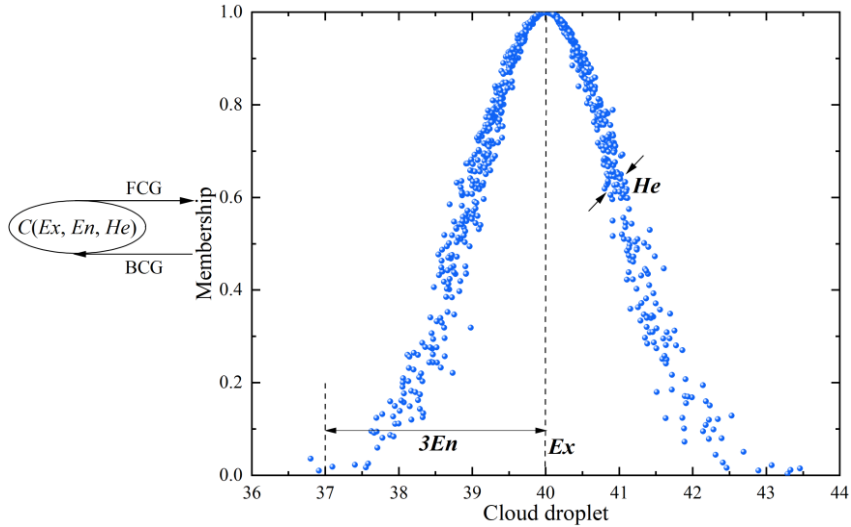

**Fig. 1.** Bidirectional cognitive transformation diagram of cloud model

**Table 1**
Correspondence between cloud model and statistical learning

| Cloud Model | Statistical learning |
|---|---|
| Expected Value | Expectation of Posterior Estimation |
| Entropy | Variance of Likelihood Distribution |
| Hyper Entropy | Hyper-parameter |
| Forward Cloud Generator | Distributed Random Number Generator |
| Backward Cloud Generator | Point Estimator |
| Cloud Droplet | Random Variable |
| Cloud Transformation | Mixed Distribution Model |
| Cloud Model | Measure Space |

In this paper, SCN-CEL is proposed to handle the problems of uncertain information and few-shot learning of rolling bearing fault diagnosis, and its overall architecture is given in Fig. 2. It should be noted that the small number of samples come from the same type of fault. The procedure of SCN-CEL consists of cloud feature extraction, cloud sampling and cloud ensemble learning.

*2.2 Cloud feature extraction*

The vibration signals of different health states of rolling bearing are denoised by the wavelet filter and normalized given by Eq. (1).

$$\widehat{x}_{p,q} = \frac{x_{p,q} - x_p^{min}}{x_p^{max} - x_p^{min}}, p = 1...m \qquad (1)$$

where *m* represents the total number of healthy states, $x_{p,q}$ is the value recorded by vibrating sensor at the *q*-th sampling point under the *p* health state. Generate lots of samples on the original vibration signals by a sliding window, as shown in Fig. 3.

Feature entropy (*En*) of cloud model overcomes the difficulty of selecting traditional entropy parameters when solving uncertainty problems, and can effectively express the fault characteristics of rolling bearings [34]. Essentially, the BCG of the one-dimensional cloud model is applied to extract the fault features (*Ex*, *En*, *He*) of rolling bearing from each sample divided in Fig.3. Therefore, this is a process of backward cloud transformation for obtaining the intension knowledge (*Ex*, *En*, *He*) of vibration signal in certain health states. The steps of feature extraction are listed as follows.

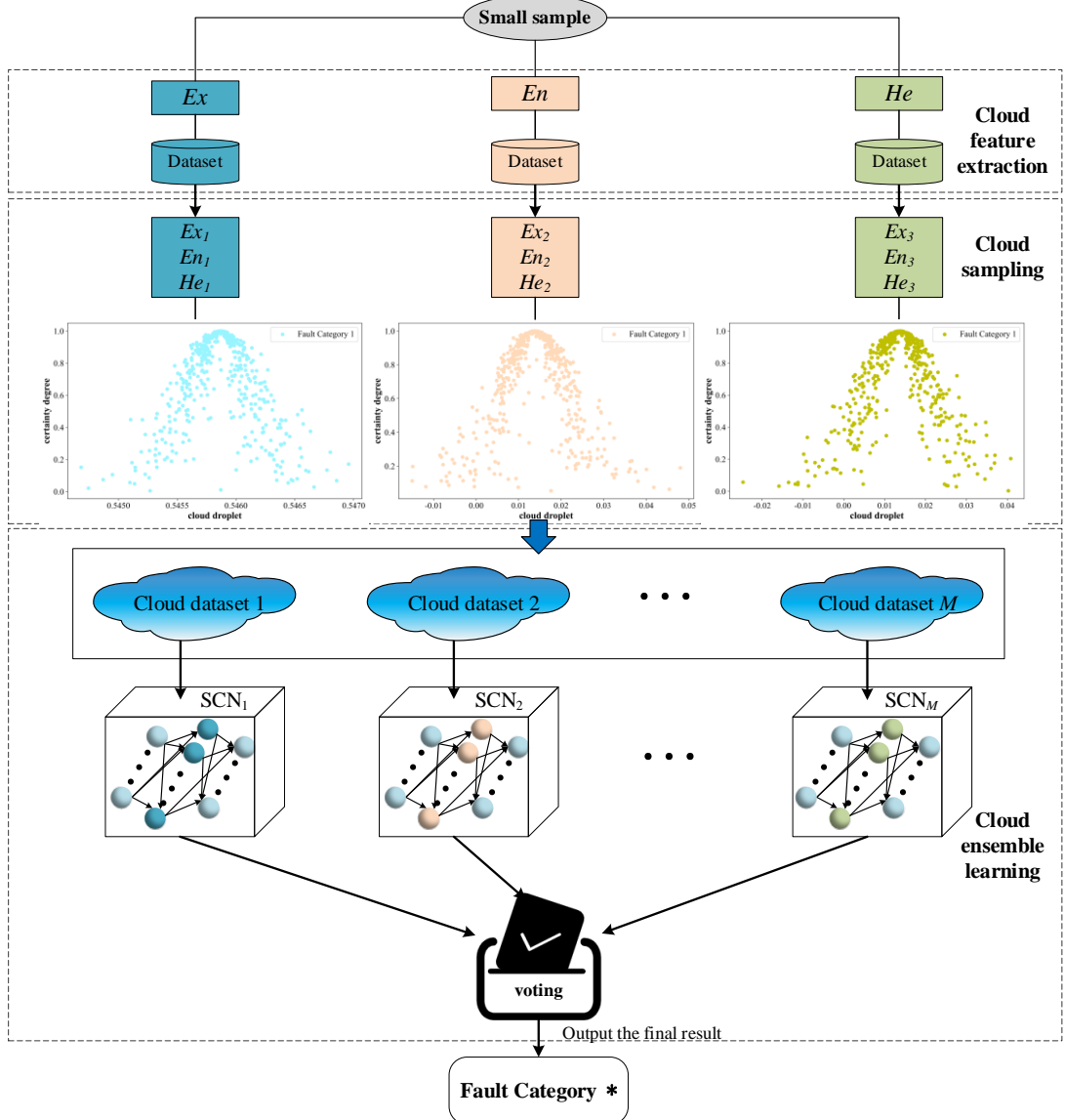

**Fig. 2**. Overall architecture of the SCN-CEL

(1) $x = [x_1, x_2, \cdots, x_{size}]$ is an original sample preprocessed as Fig. 3 where *size* is the length of the sliding window. Get its sample mean $\bar{X} = 1 \Big/ size \sum_{i=1}^{size} x_i$, first order center distance $1 \Big/ size \sum_{i=1}^{n} |x_i - \bar{X}|$ and sample variance $S^2 = 1 \Big/ ((n-1) \sum_{i=1}^{n} (x_i - \bar{X})^2)$.

(2) Calculate the expectations of cloud:

$$Ex = \bar{X} \qquad (2)$$

(3) Calculate the feature entropy of cloud:

$$En = \sqrt{\frac{\pi}{2}} \times \frac{1}{size} \sum_{i=1}^{size} |x_i - Ex| \qquad (3)$$

(4) Calculate the hyper entropy of cloud：

$$He = \sqrt{|S^2 - En^2|} \quad (4)$$

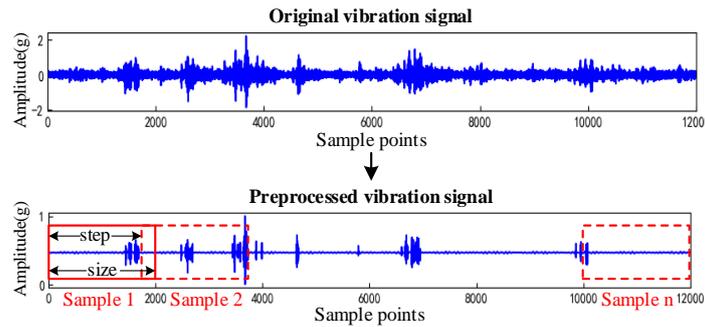

Fig. 3. Preprocessing of vibration signal of certain health state

*2.3 Cloud sampling*

Overall feature datasets are obtained by performing the aforementioned feature extraction on all fault signals and normal signals. Small training sets are composed of a small number of samples randomly selected from the same fault type in feature datasets. The samples of the same fault type are separated from small training set and *Ex, En* and *He* are calculated again for each dimensional feature. Noteworthily, *En* is the uncertainty measure of qualitative concept-randomness measure and ambiguity (the size of acceptable cloud droplets in the domain), reflecting the discreteness of clouds. *He* is a measure of uncertainty of *En*. The larger the *He*, the greater the dispersion of cloud droplets. With those *Ex*, *En* and *He*, enough cloud droplets are generated to augment small samples according to the FCG. In the process of cloud sampling, sampling parameters (*Ex*, *En*, *He*) are solved according to the existing training samples, and no manual setting of hyper-parameters is required. The above process is shown in the following steps.

(1) Suppose $S_i (i=1,2,\cdots,m)$ is a small training set under a fault category where *m* is the number of fault categories. Based on the characteristics of $S_i$, $Ex_{i,j}, En_{i,j}$ and $He_{i,j} (j=1,2,3)$ are calculated by Eqs. ((2)-(4)).

(2) Produce a Gaussian random number $En'_{i,j}$ with expectation of $En_{i,j}$ and variance of $He_{i,j}^2$, as shown in Eq. (5).

$$En'_{i,j} = NORM(En_{i,j}, He_{i,j}^2) \quad (5)$$

(3) Generate a Gaussian number $x_{i,j}^q (q=1,2,\cdots,N)$ by Eq. (6) where *N* is the sum of cloud droplets.

$$x_{i,j}^q = NORM(Ex_{i,j}, En'^{2}_{i,j}) \quad (6)$$

Its degree of certainty is $\mu_{i,j}^q = \exp\left[(x_{i,j}^q - Ex_{i,j})^2 / 2En'^{2}_{i,j}\right]$ while $\mu_{i,j}^q$ and $x_{i,j}^q$ represent the cloud droplet in the mathematic domain and are marked as $drop(x_{i,j}^q, \mu_{i,j}^q)$ ).

(4) Repeat Steps ((2)-(3)) until *N* cloud droplets are produced.

(5) Execute the aforementioned steps for all feature dimensions of different fault categories to ultimately acquire cloud sample datasets.

It is conspicuous that cloud sampling leverages the bidirectional cognitive transformation of cloud model to explore the fault information with uncertainty from raw feature distribution. Finally, the implementation procedure of cloud sampling is summarized in **Algorithm 1**.

| **Algorithm 1: Cloud Sampling** |
|---|
| Given a small training set $X = \{x_1, x_2, \cdots, x_N\}$, $x_i \in R^d$ as inputs and outputs $Y = \{y_1, y_2, \cdots, y_N\}$, $y_i \in R^m$; *m* is the number of health states of rolling bearing; *N* is the number of cloud droplets. |
| 1. Initialize one cloud sample set $CsampleX = [\ ], CsampleY = [\ ]$; |
| 2. **For** $i = 1, 2, \cdots, m$, **Do** |
| 3.   Initialize $X'_i = [\ ], Y'_i = [\ ]$; |
| 4.   Get $X_i = \{X|Y=i\}$; |
| 5.   **For** $j = 1, 2, \cdots, d$, **Do** |
| 6.     Initialize $Drops_{i,j}X = [\ ]$, $Drops_{i,j}Y = [\ ]$; |
| 7.     Calculate $Ex_{i,j}, En_{i,j}, He_{i,j}$ in $X_i{}^j$ based on Eqs. ((2)-(4)); |
| 8.     **While** $len(Drops_{i,j}) \leq N$, **Do** |
| 9.       Add *i* to $Drops_{i,j}Y$; |

| | |
|---|---|
| **10.** | Generate one cloud droplet according to $Ex_{i,j}, En_{i,j}, He_{i,j}$ and add to $Drops_{i,j}X$ based on Eqs. ((5)-(6)); |
| **11.** | **End While** |
| **12.** | Let $Drops_{i,j}X^T$ add to $X'_i$; |
| **13.** | Let $Drops_{i,j}Y^T$ add to $Y'_i$; |
| **14.** | **End For** (corresponds to **Step 5**) |
| **15.** | Add $X'_i$ to $CsampleX$ by column; |
| **16.** | Add $Y'_i$ to $CsampleY$ by column; |
| **18.** | **End For** (corresponds to **Step 3**) |
| **19. Return** $CsampleX$ and $CsampleY$ | |

*2.4 Cloud ensemble learning*

In order to comprehensively expresses the uncertainty information introduced by the cloud model and improve the generalization performance of individual SCN, cloud ensemble learning that executes cloud sampling multiple times to generate diverse cloud datasets and conducts fault diagnosis based on ensemble SCN is developed. The cloud datasets serve as the input of base classifier of ensemble SCN and maintain the diversity of base classifier. Eventually, the fault diagnosis result is predicted through a majority voting. $f_k(x)$ $(k=1,2,\cdots,K)$ is the decision function of the $k$-th SCN in SCN-CEL and $Y_j(j=1,2,\cdots,m)$ is the label of the $j$-th sample. $Num_j = number\{k|f_k(x)=C_j\}$ means the total amount of $j$-th label in all predictions from SCN, where $number(\cdot)$ indicates counting the number. The final classification results can be obtained by

$$f_{final}(x) = \arg\max_j(Num_j) \qquad (7)$$

Its pseudocode is demonstrated in **Algorithm 2**.

| **Algorithm 2: SCN-CEL** |
|---|
| Given a small training set $X = \{x_1, x_2, \cdots, x_N\}$, $x_i \in R^d$ as inputs and outputs $Y = \{y_1, y_2, \cdots, y_N\}$, $y_i \in R^m$; $m$ is the number of health states of rolling bearing; $K$ is the number of SCN. |
| **1.** Obtain $CsampleX_q$ and $CsampleY_q$ ($q=1,2,...,K$) from **Algorithm 1** using $X$ and $Y$; |
| **2. For** $q=1,2,\cdots K$, **Do** |
| **3.** Get a prediction result $Y_q^*$ of $q$-th SCN in ensemble SCN; |
| **4. End For** (corresponds to **Step 2**) |
| **5.** Calculate the number of various labels based on $Y^*$; |
| **6. Return** final fault category based on Eq. (7) |

## 3. Experimental study

In this section, the rolling bearing vibration acceleration signals from the Bearing Data Center of Case Western Reserve University [35] are used to demonstrate the effectiveness and performance of the proposed approach. As shown in Fig. 4, the test signals are acquired from the 6205-2RSJEM SKF deep groove ball bearing and the test platform comprises a 2 HP motor, a torque sensor/encoder, a dynamometer, and control electronics which is not demonstrated in that figure. The bearing faults consisting of inner ring (IR) fault, outer ring (OR) fault (its damage point is set at 3 o'clock, 6 o'clock and 12 o'clock), rolling element (RE) fault and normal state are caused by pitting corrosion using electrical discharge with these fault diameters: 0.1778 mm (7 mils), 0.3556 mm (14 mils), and 0.5334 mm (21 mils). And the vibration acceleration data are recorded under four kinds of operation conditions corresponding to the motor speeds of 1797 r/min with 0 HP load, 1772 r/min with 1 HP load, 1750 r/min with 2 HP load, and 1720 r/min with 3 HP load, with a sampling frequency of 12kHz. All experiments are run in Window 10 operating system (Intel i5-12600KF 3.69GHz CPU, 32G RAM).

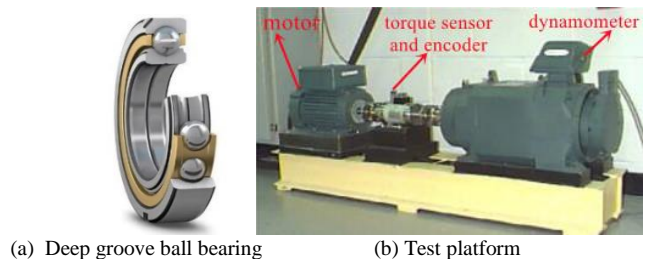

(a) Deep groove ball bearing  (b) Test platform
**Fig. 4.** Construction of test platform

*3.1 Experiment 1. Effectiveness of cloud sampling*

In this section, comparisons among the cloud sampling, kernel density estimation (KDE) based sampling [36], synthetic minority oversampling technique (SMOTE) [37], adaptive synthetic sampling approach (ADASYN) [38] and bootstrap oversampling (BOS) are carried out to explore the effectiveness and performance of cloud sampling. Aiming at addressing the problem of small fault samples which is also the problem of an unbalanced distribution of sample data, five resampling methods belonging to oversampling are utilized to augment the number of small samples. Resampling involves balancing the class distribution by either undersampling the majority class or oversampling the minority class.

Bootstrap, an import method for non-parametric estimates, is often used for statistical inference of a model. BOS is the simplest approach to generate additional minority fault samples by repeatedly randomly selecting from the existing set. Therefore, BOS is able to serve as a basic benchmark for other sampling methods. SMOTE first calculates the *K*-neighbors of each sample in the minority class, randomly selects a subset of these neighbors based on the degree of imbalance, and then constructs a synthetic sample using various mathematical formulas applied to the selected subset. ADASYN, known as a popular variant of SMOTE, is an adaptive algorithm to generate samples. Based on the estimated probability density distribution of small samples, the KDE-based oversampling method can generate more samples.

The first sub-experiment data consists of two-dimensional data points involving RE faults (10 groups) and Normal signals (60 groups) with fault size 7 mils from the drive end bearing when the motor worked at 0 HP load. The aforementioned five methods are utilized to oversample the fault signals until the relatively balanced data distribution is achieved. That is presented in Fig. 5.

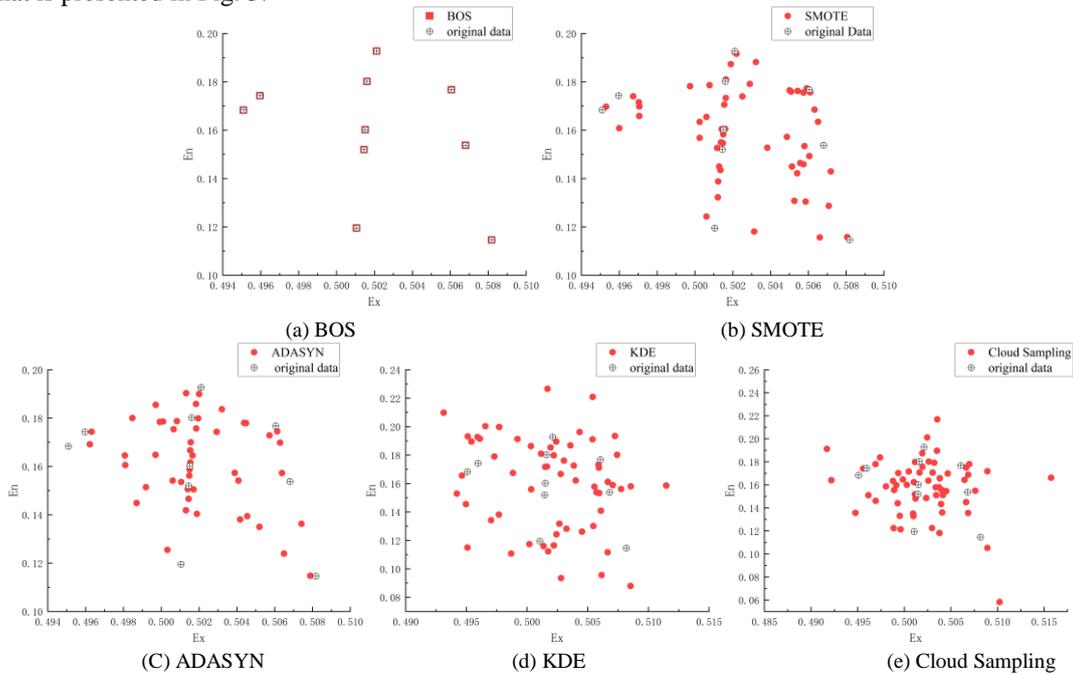

(a) BOS  (b) SMOTE
(C) ADASYN  (d) KDE  (e) Cloud Sampling

**Fig 5.** Resampled data based on various methods.

The new points are created by simply resampling the existing fault samples in BOS. SMOTE fabricates new samples between different samples. ADASYN is similar to SMOTE, but it pays more attention to the border of interclass samples. The samples generated by KDE are relatively divergent, a small number of which are outside the minority class intuitively. For KDE, cloud sampling brings the generated data to gather inward, which suggests it's based on normal distribution.

**Table 2**
Detailed information of the feature datasets

| No. | Health state | Fault diameter (mils) | Groups | Motor load |
|---|---|---|---|---|
| 1 | Normal | - | 70 | 0 HP |
|  | IR fault | 7 | 10 |  |
|  | OR@3 fault | 7 | 10 |  |
|  | OR@6 fault | 7 | 10 |  |
|  | OR@12 fault | 7 | 10 |  |
|  | RE fault | 7 | 10 |  |
| 2 | Normal | - | 70 | 1 HP |
|  | IR fault | 7 | 10 |  |
|  | OR@3 fault | 7 | 10 |  |
|  | OR@6 fault | 7 | 10 |  |
|  | OR@12 fault | 7 | 10 |  |
|  | RE fault | 7 | 10 |  |
| 3 | Normal | - | 70 | 0 HP |
|   | IR fault | 21 | 10 |   |
|   | OR@3 fault | 21 | 10 |   |
|   | OR@6 fault | 21 | 10 |   |
|   | OR@12 fault | 21 | 10 |   |
|   | RE fault | 21 | 10 |   |
|   | Normal | - | 70 |   |
|   | IR fault | 21 | 10 |   |
| 4 | OR@3 fault | 21 | 10 | 1 HP |
|   | OR@6 fault | 21 | 10 |   |
|   | OR@12 fault | 21 | 10 |   |
|   | RE fault | 21 | 10 |   |
|   | Normal | - | 70 |   |
| 5 | IR fault | 7,14,21 | 30 | 2 HP |
|   | OR@6 fault | 7,14,21 | 30 |   |
|   | RE fault | 7,14,21 | 30 |   |

A single SCN is adapted as the classifier for bearing fault detection to further verify the effectiveness of cloud sampling and analyze its differences with other methods. The $T_{max}$, $L_{max}$ and $e$ of SCN are 100, 150 and 0.1, respectively. All samples are intercepted by sliding window and preprocessed, as described in Section 2.2. As mentioned above, each group of feature vectors is obtained by backward cloud generator. The detailed information of prepared feature datasets which contain small samples for each fault is shown in Table 2. "OR@3" indicates that the damaged point is located at the 3 o'clock direction of outer ring. The samples are gathered by a sliding window with a length of 500 and a step size of 200. The number of small samples is set to 10 and the number of normal samples is set to 70, which constitutes an unbalanced data distribution. And the number of each health state for testing is set to 30 while the training samples of faults are generated on a 7:3 scale by mentioned sampling methods except ADASYN. It is worth mentioning that in order to highlight the validity of our expanded data, the training samples are all fabricated samples instead of original small samples.

Fig. 6 shows the visualization of No.1, No.2 and No.3 dataset, with the left figures illustrating their distribution of cloud features and the right figures displaying their projection on the *Ex-En* plane. Firstly, the normal sample points are remote from the fault samples because of their relatively small amplitude. OR@12 fault has a significant impact on the vibration magnitude and does not deteriorate the vibration stability when fault severity is the same. The vibration of the rolling bearing affected by the rolling element is more chaotic and more intense while the OR@12 fault makes the vibration more intense, but more stable. As shown in Fig. 6 (a) and (c), it is known that the rolling element fault has slight impact on the vibration amplitude. Simultaneously, the vibration signal is irregular, which could be caused by the random change of the contact point between the ball and the raceway and the role of lubricating oil, and IR fault causes the amplitude to greatly increase.

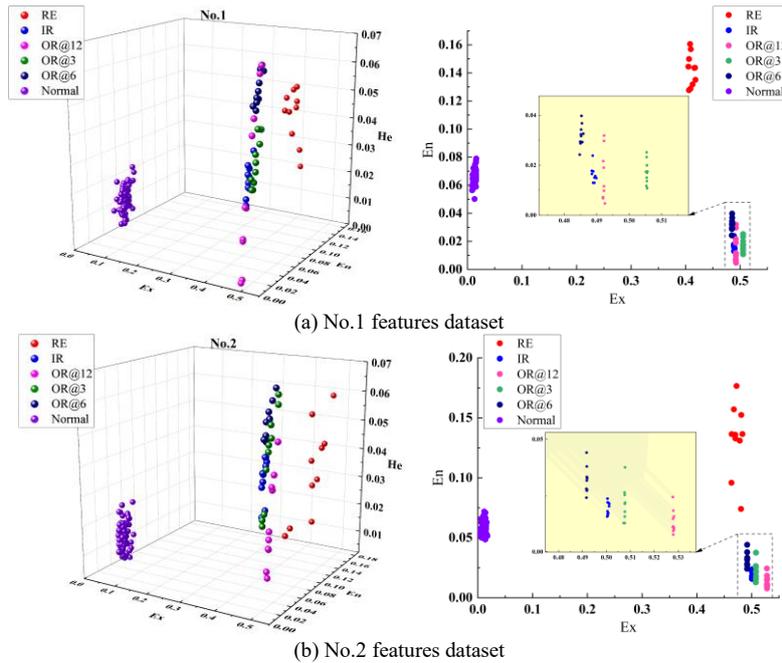

(a) No.1 features dataset

(b) No.2 features dataset

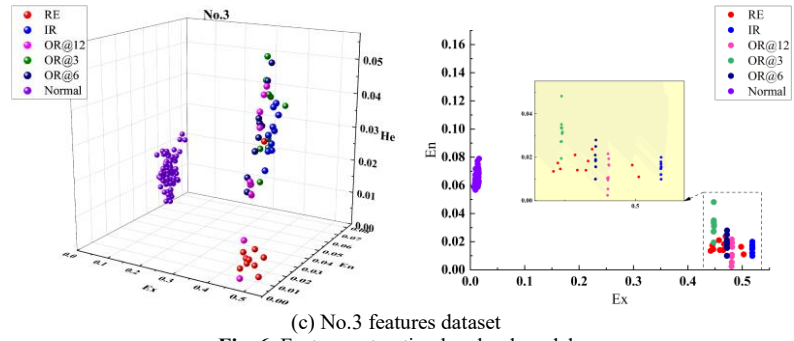
(c) No.3 features dataset
**Fig. 6.** Feature extraction by cloud model

Following the acquisition of training samples with diverse sampling methods, the average results presented in Fig. 7 and Fig. 8 are obtained by repeating the experiment 100 times. Compared to our benchmark oversampling method BOS, the testing accuracy of cloud sampling for fault diagnosis is significantly higher, which proves that our proposed method is very effective in solving the issue of fault diagnosis with small samples. It can be seen that there is a serious overfitting in BOS which does not generate new samples from Fig.5, while ADASYN and SMOTE method also show a slight overfitting phenomenon. Compared to ADASYN, SMOTE, and KDE, cloud sampling is slightly less accurate than KDE in the first four datasets, but more accurate than other methods. However, the training time of the SCN is shorter than that of KDE. The new samples generated by ADASYN and SMOTE are all within or on the boundary of small samples, which makes the distribution of the new samples less complicated. That is why SCN with KDE and cloud sampling have longer training time than other methods. In the experiment of the No.5 dataset, cloud sampling has a higher testing accuracy than KDE but is slightly inferior to ADASYN and SMOTE. The testing accuracy of BOS, SMOTE, and ADASYN fluctuates greatly, whereas that of cloud sampling remains relatively stable.

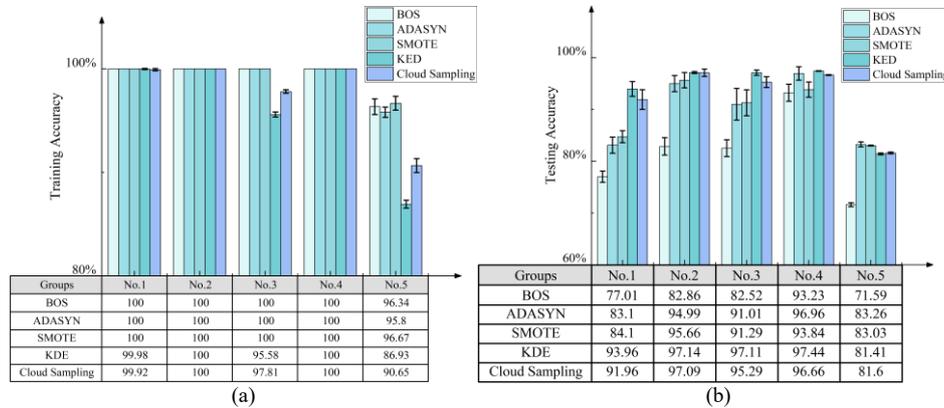

**Fig. 7.** (a) Training accuracy, (b) Testing accuracy of SCN with different sampling methods in dataset of each group.

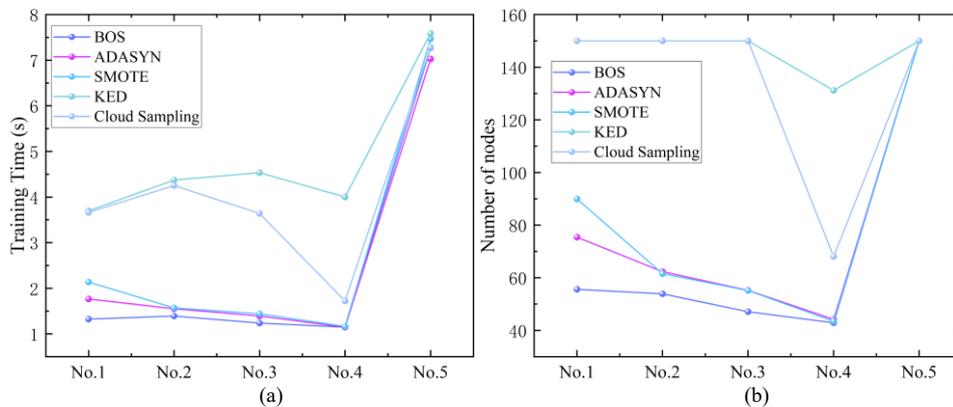

**Fig. 8.** (a) Training time, (b) The number of nodes of SCN with different sampling methods in dataset of each group.

*3.2 Experiment 2. Diagnostic accuracy comparisons of existing fault diagnosis methods*

This experiment compares SCN-CEL with several algorithms listed in Table 3 that have performed well in bearings fault diagnosis, aiming to showcase the superiority of the proposed method with small fault samples. It is worth mentioning that in order to reflect the important role of ensemble learning and cloud model in solving the problem of small samples for fault diagnosis, we propose C-SCN which leverages cloud model for feature extraction but does not rely on ensemble learning for fault type identification to make a comparison.

Table 3
Bearing fault diagnosis methods for comparison

| Index | Approach | Methodology | Abbreviation |
|---|---|---|---|
| 1 | Zhang et al. [17] | Deep convolutional neural Networks with wide first-layer kernels | WDCNN |
| 2 | Chen et al.[20] | Multi-scale convolutional neural network and long short-term memory | MCNN-LSTM |
| 3 | Aburakhia et al.[39] | Bayesian optimization-based random forest algorithm | Bayesian-RF |
| 4 | Li et al.[15] | Random vector functional link neural network | RVFLN |

a. Experimental data

The overall datasets are still generated by sliding the window with a length of 1024 and a step size of 200. Then we get the original training samples and testing samples. The size of testing samples for each healthy state is 30. The detailed information of datasets is listed in Table 4. As for SCN-CEL, the input of 70 fault samples for training is generated by cloud sampling without containing the original cloud features data.

b. Parameter settings

The sigmoid function is designated as the activation function for RVFLN, C-SCN, and SCN-CEL. The $T_{max}$, $L_{max}$ and $e$ for C-SCN and SCN-CEL are 100, 100 and 0.1, respectively. The number of SCN for SCN-CEL is set to 6, indicating cloud sampling is conducted for 6 times. The optimal number of hidden nodes of RVFLN are set through grid search. The training epochs and batch size of WDCNN and MCNN-LSTM are 400 and 40, and their learning rate = 0.0006. The $m$ and $k$ of Bayesian-RF are set to 1 and 3.

c. Experimental results and analysis

The results presented in Fig. 9 substantiate that the proposed method outperforms other methods in terms of diagnostic performance of rolling bearings under diverse operational conditions with small samples, as evidenced by an average overall accuracy of 50 trials. From the results of C-SCN, it can be seen that proposed method for feature extraction through backward cloud generator is so functional. At the same time, the great generalization performance of SCN is demonstrated. Compared to RVFLN, which requires the selection of hidden nodes, setting the number of SCN for C-SCN is a simple parameter adjustment. SCN-CEL has not only higher accuracy than C-SCN, but also less fluctuation in accuracy. WDCNN and MCNN-LSTM, as fault diagnosis methods based on deep learning, have many drawbacks when dealing with a few fault samples. MCNN-LSTM does not achieve 90% testing accuracy described in the original paper due to our insufficient training samples and its undersampling of raw signals, which may result in loss of feature information. As the fault diameter increases, the prediction accuracy of some methods decreases, but the accuracy of SCN-CEL declines slightly. It is worth noting that the training samples of SCN-CEL only contain cloud samples without raw feature samples, while SCN-CEL can achieve such a high accuracy for fault diagnosis under condition of unbalanced training and small number of fault samples.

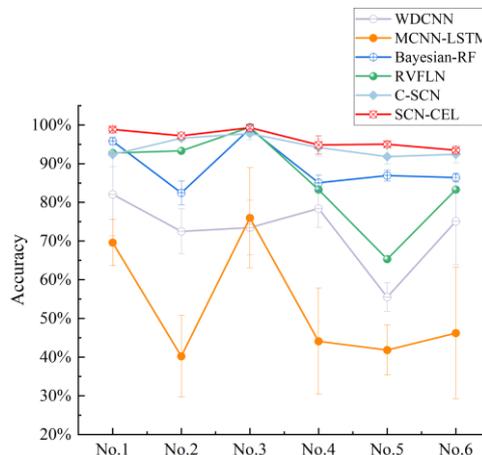

**Fig. 9.** Diagnostic accuracy comparisons of existing fault diagnosis methods

## 3.3 Experiment 3. Effect of the number of training samples

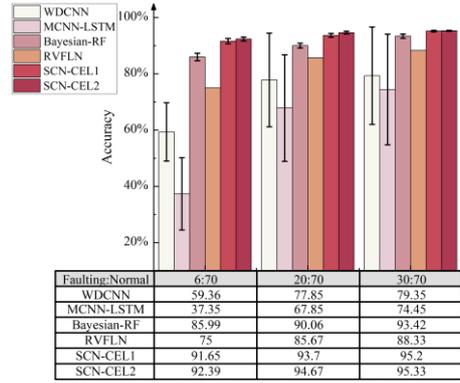

| Faulting:Normal | 6:70 | 20:70 | 30:70 |
|---|---|---|---|
| WDCNN | 59.36 | 77.85 | 79.35 |
| MCNN-LSTM | 37.35 | 67.85 | 74.45 |
| Bayesian-RF | 85.99 | 90.06 | 93.42 |
| RVFLN | 75 | 85.67 | 88.33 |
| SCN-CEL1 | 91.65 | 93.7 | 95.2 |
| SCN-CEL2 | 92.39 | 94.67 | 95.33 |

**Fig. 10.** Accuracy comparisons with existing fault diagnosis methods in various number of training samples

The testing accuracy of SCN-CEL is the lowest in the No.6 dataset, as indicated in Fig.9. The goal of cloud sampling is to generate more training samples based on a small number of existing fault samples. Thus, comparing SCN-CEL with other methods involved in Table 3, the effect of training samples number is explored in this experiment. Unlike Experiment 2, the training datasets of SCN-CEL are divided into two types: one is filled with cloud samples and the other is a combination of cloud samples and existing small samples, referred to as "SCN-CEL1" and SCN-CEL2", respectively. The parameter configuration of this experiment is the same as Experiment 2. Fault diameter, health state and motor load settings of this experiment are identical to those of the No.6 dataset in Table 4. However, the number of samples for each fault type and normal samples is divided into three categories: 6:70, 20:70, 30:70. The fault samples of each type are split based on its previous type and testing samples are consistent across different types. As for SCN-CEL, the amount of generated training samples are 70, 140, and 210, respectively.

**Table 4**
Detailed information of original training datasets

| No. | Health state | Fault diameter (mils) | Groups | Motor load |
|---|---|---|---|---|
| 1 | Normal | - | 70 | 1 HP |
|   | IR fault | 7 | 10 |  |
|   | OR@3 fault | 7 | 10 |  |
|   | OR@6 fault | 7 | 10 |  |
|   | OR@12 fault | 7 | 10 |  |
|   | RE fault | 7 | 10 |  |
| 2 | Normal | - | 70 | 1 HP |
|   | IR fault | 21 | 10 |  |
|   | OR@3 fault | 21 | 10 |  |
|   | OR@6 fault | 21 | 10 |  |
|   | OR@12 fault | 21 | 10 |  |
|   | RE fault | 21 | 10 |  |
| 3 | Normal | - | 70 | 2 HP |
|   | IR fault | 7 | 10 |  |
|   | OR@3 fault | 7 | 10 |  |
|   | OR@6 fault | 7 | 10 |  |
|   | OR@12 fault | 7 | 10 |  |
|   | RE fault | 7 | 10 |  |
| 4 | Normal | - | 70 | 2 HP |
|   | IR fault | 21 | 10 |  |
|   | OR@3 fault | 21 | 10 |  |
|   | OR@6 fault | 21 | 10 |  |
|   | OR@12 fault | 21 | 10 |  |
|   | RE fault | 21 | 10 |  |
| 5 | Normal | - | 70 | 1 HP |
|   | IR fault | 7,14,21 | 30 |  |
|   | OR@6fault | 7,14,21 | 30 |  |
|   | RE fault | 7,14,21 | 30 |  |
| 6 | Normal | - | 70 | 3HP |
|   | IR fault | 7,14,21 | 30 |  |
|   | OR@6fault | 7,14,21 | 30 |  |
|   | RE fault | 7,14,21 | 30 |  |

The average results of our 50 trials is shown in Fig. 10. The proposed method achieves the best performance with the various number of unbalanced training samples. The overall accuracy of SCN-CEL2 is the real accuracy that our method can achieve. With the increase in the number of training samples, the test accuracy of all methods is improved, which suggests that the number of training samples has a great influence on the methods based on deep learning. The total average accuracy of SCN-CEL1 and SCN-CEL2 both exceed 90% with a minimal variance in these three datasets. With the number of training samples increasing, the accuracy variance of WDCNN and MCNN-LSTM grows larger while that of the proposed method diminishes. In a nutshell, our approach is capable of accurately and steadily detecting bearing faults with a few of samples by generating additional training samples tailored to different practical requirements.

## 4. Conclusions

To address the problem of fault diagnosis for rolling bearings in the few shot scenarios, a new approach named SCN-CEL is proposed by combining the cloud model and the stochastic configuration networks (SCNs). In order to further improve the generalization performance of the classifier, this approach applies ensemble learning based on cloud sampling to SCNs, which can exactly multiple times use cloud samples generated through backward cloud generator to fully reflect the uncertainty introduced by the cloud model. Three different types of experiments are conducted using the CWRU bearing data set. Experiment 1 verifies the proposed cloud sampling which is used to oversample the feature datasets of original fault samples is very effective. The SCNs with cloud sampling can achieve a satisfactory testing accuracy of fault diagnosis in various working conditions and fault severity for rolling bearing. Compared to existing methods of fault diagnosis, SCN-CEL shows better predictive accuracy with small fault samples, and the efficiency of feature extraction through forward cloud generator has been proved in Experiment 2. Considering the effect of the number of training samples, Experiment 3 is conducted to confirm the highest accuracy of proposed approach with a small variance. These experiments fully demonstrate the great superiority of this method over other methods mentioned in this paper.

For future work, the distribution of cloud features is worthy of further study to improve cloud sampling theory and more effectively capture feature information from the original vibration signals. And the compound fault is a prevailing issue in reality, thus it is worthwhile to explore the utilization of cloud models for its resolution.

## Acknowledgments

This work was supported in part by the National Natural Science Foundation of China under Grant 61973306 and the Natural Science Foundation of Jiangsu Province under Grant BK20200086.